\documentclass[a4paper,twoside]{article}
\usepackage[table]{xcolor}
\usepackage{floatrow}
\DeclareFloatFont{tiny}{\tiny}
\usepackage{graphicx}
\usepackage{subfigure}
\usepackage{amsmath}
\usepackage{rotating}
\usepackage{color, soul}
\usepackage[inkscapeformat=png]{svg}
\usepackage{svg}
\usepackage{hyperref}
\usepackage{tabularray}
\UseTblrLibrary{booktabs}
\usepackage{graphicx}%
\usepackage{lineno}
\usepackage{floatrow}
\usepackage{tabularx,booktabs, longtable}
\usepackage{amsmath,amssymb,amsfonts}%
\usepackage{amsthm}%
\usepackage{mathrsfs}%
\usepackage[title]{appendix}%
\usepackage{textcomp}%
\usepackage{manyfoot}%
\usepackage{booktabs}%
\usepackage{longtable}
\usepackage{algorithmicx}%
\usepackage{algpseudocode}%
\usepackage{listings}%
\usepackage{tabularray}
\usepackage{hyperref}
\usepackage{framed,multirow}
\usepackage{comment}
\DefTblrTemplate{contfoot-text}{default}{}
\DefTblrTemplate{conthead-text}{default}{}
\DefTblrTemplate{caption}{default}{}
\DefTblrTemplate{conthead}{default}{}
\DefTblrTemplate{capcont}{default}{}
\usepackage{epsfig}
\usepackage{subcaption}
\usepackage{calc}
\usepackage{natbib}
\usepackage{xcolor}
\usepackage{tabularx}
\usepackage{amssymb}
\usepackage{amstext}
\usepackage{anyfontsize}
\usepackage{amsmath}
\usepackage{graphicx}
\usepackage{amsthm}
\usepackage{multicol}
\usepackage{multirow}
\usepackage{pslatex}
\usepackage[super]{nth}
\let\bibhang\relax
\usepackage{apalike}
\usepackage{algorithm2e}
\usepackage[bottom]{footmisc}
\usepackage{SCITEPRESS}     

\begin{document}

\title{Transformer-based Video Saliency Prediction with High Temporal Dimension Decoding}

\author{\authorname{Morteza Moradi, Simone Palazzo, Concetto Spampinato} 
\affiliation{PeRCeiVe Lab, University of Catania, Italy}
\email{morteza.moradi@phd.unict.it}
}

\keywords{Video Saliency Prediction, Gaze Prediction, Visual Attention, Spatio-temporal Transformer}

\abstract{In recent years, finding an effective and efficient strategy for exploiting spatial and temporal information has been a hot research topic in video saliency prediction (VSP). With the emergence of spatio-temporal transformers, the weakness of the prior strategies, e.g., 3D convolutional networks and LSTM-based networks, for capturing long-range dependencies has been effectively compensated. While VSP has drawn benefits from spatio-temporal transformers, finding the most effective way for aggregating temporal features is still challenging. To address this concern, we propose a transformer-based video saliency prediction approach with high temporal dimension decoding network (THTD-Net). This strategy accounts for the lack of complex hierarchical interactions between features that are extracted from the transformer-based spatio-temporal encoder: in particular, it does not require multiple decoders and aims at gradually reducing temporal features’ dimensions in the decoder. This decoder-based architecture yields comparable performance to multi-branch and over-complicated models on common benchmarks such as DHF1K, UCF-sports and Hollywood-2.}

\onecolumn \maketitle \normalsize \setcounter{footnote}{0} \vfill

\section{\uppercase{Introduction}}

Saliency prediction aims to model the behavior of the human visual system to predict fixation points when freely observing a visual scene. Recently, saliency prediction has gained great importance in different computer vision tasks, including medical analysis \citep{arun2021assessing}, gaze estimation \citep{DBLP:conf/ijcai/AbawiWW21}, video compression \citep{lyudvichenko2017semiautomatic}, object segmentation \citep{chen2020contour}, to name a few.  When it comes to video saliency prediction (VSP) \citep{wang2019revisiting} the problem becomes more challenging because not only should spatial features be taken into account, but also long-range temporal dependencies among frames. In this regard, different approaches for capturing spatio-temporal features in VSP have been introduced. One of the most common strategies in this domain is to capture spatial and temporal features in two separate parallel branches and integrating these features to form the spatio-temporal features for facilitating the video saliency prediction process \citep{zhang2020spatial}. Another technique for extracting and exploiting spatio-temporal information involves LSTM networks \citep{fang2020devsnet}, and temporal information is extracted based on processing a sequence of spatial features, without synchronizing temporal and spatial features.

To address the weakness of LSTM-based models, 3D convolution-based VSP models \citep{HD2S} improved the performance of the video saliency prediction models to a great extent by aggregating the spatial and temporal features at the same time using 3D convolutional layers. On the other hand, for capturing long-range temporal dependencies among consecutive frames, 3D convolutional networks fall short since they extract spatio-temporal information within a fixed space-time local window. To alleviate the aforementioned problem, the emergence of spatio-temporal vision transformers \citep{li2023uniformer} has revolutionized video understanding research. Since transformers process patches at a global level, they are capable of capturing and comparing features among spatial and temporal dimensions at long distances, in contrast to 3D convolution-based networks, which instead rely on aggregation and downsampling. 

However, it is of high importance to select an effective strategy for employing and reducing the temporal dimension of the features that are provided by the encoder to the decoding stage. For instance, in \citet{zhou2023transformer}, a computationally expensive VSP model is introduced in which the temporal dimension of the spatio-temporal features that are provided by the encoder is reduced and then these temporally downsampled features feed the decoder. Such an approach can deprive the network's decoder of long-range dependencies that are extracted in the transformer-based encoder's multi-level features. A pure-transformer based architecture is proposed in \citet{DBLP:journals/tcsv/MaSRZL22}. This network employs attention mechanism in order to extract spatio-temporal dependencies between the input past frames and the target future frame. Plus, a cross-attention guidance block (CAGB) is designed and employed in the decoder that aims to aggregate the extracted multi-level representations from the encoder.

In light of the aforementioned observation, a transformer-based video saliency prediction model (THTD-Net) is presented in this paper. The underlying idea for this model is to introduce a lighter architecture that takes benefit of the whole temporal information in the decoding stage. In other words, the temporal dimensions of features provided by the transformer-based encoder is not reduced before being fed to decoder. We found that this expedient compensates the need for attention module in the decoding stage or multiple decoding branches, yielding a model with fewer parameters compared to \citet{zhou2023transformer} (the current best-performing model on the DHF1K benchmark) while yielding performance that are on par or better than the state of the art.        

Ablation studies reveal that, in order to effectively decode spatio-temporal features with high temporal dimension, it is important to employ a comparably longer decoder in terms of the number of the layers, as well as to avoid reducing the features' temporal resolution too quickly in the decoding stage. Hence, we downsample the temporal dimension of the spatio-temporal features only in some of the convolutional layers in the decoder, often leaving that dimension unchanged in consecutive layers. Nonetheless, at the same time, excessively increasing the number of decoding layers does not improve the performance, but increases the number of the model parameters and may lead to overfitting.

\section{\uppercase{Related Work}}

The main challenge of VSP is considering both spatial and temporal information for predicting the focus of visual attention. In this regard, besides extracting spatial features on regions of interest in a specific frame, the relationship between consecutive frames to track changes in objects’ locations, their shapes and relation with the surrounding context should be taken into account. A wide variety of methods have been proposed to capture and integrate such spatio-temporal features, aiming at improving the prediction accuracy while preserving computational and memory efficiency.
UNISAL \citep{droste2020unified}, in the form of an encoder-RNN-decoder architecture- predicts visual saliency both for videos and images in an integrated framework. In this work, the diverse domain shifts between video datasets and between video and image data has been addressed by specific domain adaptation techniques. Also, to extract temporal features both for video data and even static images, a Bypass-RNN module is introduced and its decoding stage is followed by a domain-adaptive smoothing component.
Another work dealing with domain adaptation is HD2S \citep{HD2S}: instead of directly generating a saliency map, multiple intermediate maps based on features obtained in different abstraction levels are produced and integrated to yield the final output. To make the model generalizable and dataset-agnostic, multi-scale domain adaption is accompanied with gradient reversal layers. 

A recent trend in the VSP area is designing multimodal models leveraging audio data to improve the accuracy of the prediction. TSFP-Net \citep{chang2021temporal} is a 3D fully-convolutional encoder-decoder model with one visual and one audio branch. The encoder part uses S3D \citep{xie2018rethinking} as the backbone for spatio-temporal features extraction, while hierarchical decoding takes into account the influence of those features in various scales and at different levels. Another multimodal encoder-decoder architecture is AViNet \citep{jain2021vinet}, that integrates visual features from ViNet (which uses S3D to encode spatio-temporal features) and audio features from SoundNet \citep{DBLP:conf/nips/AytarVT16}.

In the last few years, several VSP methods leveraging attention mechanisms have been proposed because of their property of identifying the most important and relevant information (e.g., objects), within a frame, rather than less important features (e.g., background). The SalSAC \citep{wu2020salsac} architecture is designed in the form of a CNN-LSTM-Attention network. The encoder aims to 
extract spatial information from frames. Between the encoder and decoder parts, a shuffled attention module is employed and relations between features obtained from neighboring frames are calculated through a correlation calculation layer. Then, the temporal information is captured by using correlation-based ConvLSTM modules. 

STSANet \citep{wang2021spatio} aims to remedy the weakness of 3D convolutional-based networks in capturing long-range temporal relations among video frames by embedding spatio-temporal self-attention modules into 3D convolutional network with the aim of managing temporal dependencies. The spatio-temporal features that are obtained from different levels of the network are transferred to the attention multi-scale fusion module while semantic relationships among features are considered.
MSFF-Net \citep{zhang2023multi} introduces two new modules to common architectures. Firstly, the features extracted through S3D-based encoder are sent to two frame-wise attention modules for learning temporal information. Then, the BiTSFP module is introduced, including two attention-based fusion pipelines, for bi-directional fusion that aims to add the flow of shallow location information on the basis of the previous flow of deep semantic information. One of them works on spatial features and is used to integrate neighbor information; the other incorporates channel weights of features for more efficient integration.

Following the undeniable advantages of vision transformers \citep{DBLP:conf/iclr/DosovitskiyB0WZ21} in capturing spatio-temporal relationships, several VSP works have been integrating them for saliency prediction such as TMFI-Net \citep{zhou2023transformer}. TMFI-Net employs Video Swin Transformer \citep{liu2022video} as the backbone that provides multi-level spatio-temporal features with rich contextual cues. Each decoder branch uses a multi-dimensional attention module to decode weighted features. Finally, the intermediate saliency maps of decoder branchs are combined and the final saliency map is constructed by applying a 3D convolutional layer. Another work \citep{DBLP:journals/corr/abs-2309-08220} tackles video saliency prediction and detection using transformer in a unified framework. Transformers have also been applied to video saliency forecasting, i.e., the prediction of attention regions in consecutive future frames \citep{DBLP:journals/tcsv/MaSRZL22}.

Taking into account the important role of temporal information in the decoding stage, we design a model that decodes spatio-temporal features, as provided by a transformer-based encoder, with maximum temporal dimension, while simplifying the overall model complexity. For instance, TMFI-Net employs four attention modules in each of multiple decoder branches, increasing the number of model parameters and introducing a strong temporal aggregation due to feature downsampling. Our model, instead, is designed as a single-decoder architecture, with more but simpler decoder layers, keeping the number of parameters lower than TMFI-Net. 

\section{METHOD}

\subsection{Model Architecture}
\begin{figure*}[t]
    \centering
    \includegraphics[width=0.8\textwidth]{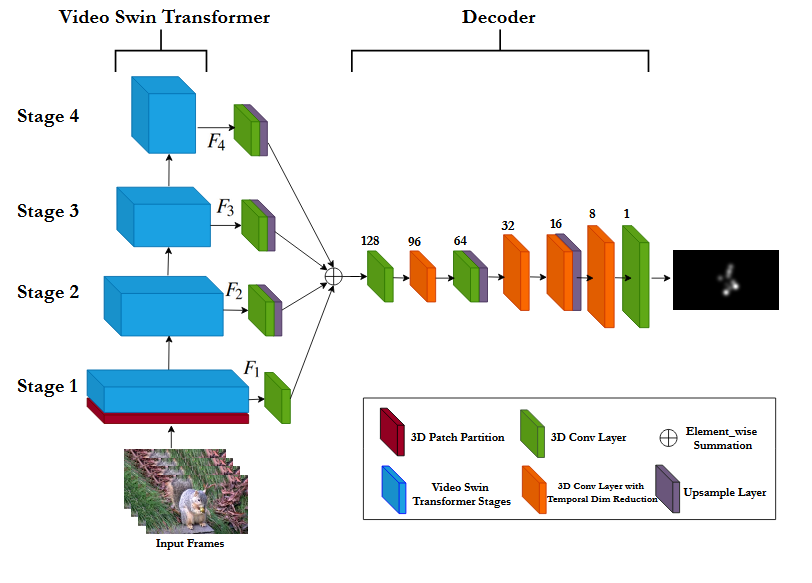}
    \caption{Overview of the proposed video saliency prediction model. The output channel dimensions of the decoder layers are reported in the figure.}
    \label{fig1}
\end{figure*}

In this section the architectural design of the proposed model is explained. Since capturing temporal dependencies is critical for video saliency prediction, and also considering the capabilities of spatio-temporal transformers for this task, THTD-Net employs the small version of Video Swin Transformer \cite{liu2022video} as the encoder. Based on the benefits of using multi-level encoder features for VSP \cite{DBLP:journals/tcsv/MaSRZL22,wang2021spatio}, Video Swin Transformer is capable to provide hierarchical multi-level features with different levels of semantic complexity at each stage. Video Swin Transformer is composed of four stages, each followed by a patch merging layer that downsamples the spatial resolution by a factor of 2, while concatenating and re-projecting features from groups of \(2\times 2\) spatially neighboring patches. The encoder network receives a video clip represented as a \(T\times H\times W\times 3\) tensor, where \(T\) is the number of consecutive frames, \(H\) and \(W\) are the height and width of each frame, respectively, and 3 is the number of color channels. Each 3D patch of size \(2 \times 4 \times 4 \times 3\) is considered as a token: the input clip is thus firstly partitioned to non-overlapping 3D patches, yielding a set of \(\frac{T}{2} \times \frac{H}{4} \times \frac{W}{4}\) tokens, each being a 96-dimensional feature. Then, a linear embedding layer projects the features of each token to dimensionality \(C\). The dimensions of the hierarchical multi-level embeddings for each stage of the encoder can be calculated as:

\begin{equation}
F_i \in \mathbb{R}^{(2^{i-1} \times C)\times \frac{T}{2}\times \frac{H}{2^{i+1}}\times \frac{W}{2^{i+1}}} 
\label{eq:F}
\end{equation} 
where \(i\) is the stage index, and \(C = 96\) is the original token dimension.

Since one of the main motivations of this work is to uncover the effectiveness of using features with high temporal dimensions for VSP, in the THTD-Net architecture the temporal dimension of the multi-level features that are provided by the encoding stages remains unaltered before feeding the decoder, in order to provide temporally-rich information for the decoding stage. 

Since the outputs of each stage have different feature dimensions (spatial resolution goes from \(\frac{H}{4} \times \frac{W}{4}\) to \(\frac{H}{32} \times \frac{W}{32}\), while the number of channels goes from $C$ to $8C$, scaling by powers of 2; the temporal dimension is unchanged and equal to $\frac{T}{2}$), 3D convolutional layers are employed after each encoder’s stage, to enable feature fusion for further decoding. In detail, a 3D convolutional layer with a $1\times 1\times 1$ kernel size is applied at each stage, providing a uniform number of feature maps of $2C$ as output; also, an upsampling layer is used at the second, third and fourth stages to bring the spatial resolution to \(\frac{H}{4} \times \frac{W}{4}\), that is the same as the first stage. 
Afterward, the spatio-temporal features from different stages are fused through element-wise summation.  

Another goal of the proposed approach is to explore the effectiveness of gradually reducing the temporal dimension of the features in the decoding stage and also employing more layers in the decoder comparing with \cite{zhou2023transformer}, in order to let the model decode spatio-temporal features more effectively. This strategy provides the opportunity of reducing the temporal resolution more gradually, avoiding the abrupt loss of information and providing rich information in each decoding stage.

The decoder thus consists of a sequence of 3D convolutional layers: kernel size for these layers is either \(2 \times 3 \times 3\) (for layers that reduce the temporal dimension; in this case, a stride of \(2 \times 1 \times 1\) is applied) or \(1 \times 3 \times 3\) (for layers that do not reduce the temporal dimension). A visualization of the model architecture is reported in Figure ~\ref{fig1}. Spatial upsampling is performed by doubling height and width of the feature maps, and the number of channels is gradually reduced from 192 to 1. We utilize a ReLU activation after each 3D convolutional layer in the decoder, except for the last layer, for which a sigmoid activation is employed.  

\subsection{Training Objective}

The model receives a video clip with \(T\) consecutive frames with the goal of predicting the saliency map on the last frame of the clip. Therefore, the training procedure is supervised by the last frame’s ground truth of the input clip. We followed the strategy introduced in \cite{zhou2023transformer} for defining the loss function as follows:

\begin{equation}
\mathcal{L}(S,G) = \mathcal{L}_\text{CC}(S,G) + \mathcal{L}_\text{KL}(S,G)
\label{eq:m}
\end{equation} 

where \(S\) is the predicted saliency map and \(G\) is the ground truth. \(\mathcal{L}_\text{CC}\) and \(\mathcal{L}_\text{KL}\) are Linear Correlation Coefficient loss and Kullback-Leibler Divergence loss, respectively. \(\mathcal{L}_\text{CC}\) calculates the linear relationships between two random variables as follows:

\begin{align}
\mathcal{L}_\text{CC}(S,G) = - \frac{\text{cov}(S,G)}{\rho(S) \rho(G)}
\end{align} 

where \(\rho(\cdot)\) is the standard deviation operator and \(\text{cov}(S,G)\) is the covariance of \(S\) and \(G\).

\(\mathcal{L}_\text{KL}\) loss aims to measure the distance between two
probability distributions and can be calculated as follows:
\begin{align}
\mathcal{L}_\text{KL}(S,G) = \sum\limits_{x} G(x) \log \frac{G(x)}{S(x)}
\end{align}
where \(x\) represents the coordinate indices.

\section{EXPERIMENTS}

\subsection{Datasets}

In order to evaluate the performance of the THTD-Net, we conduct extensive
experiments on the following commonly used datasets:
DHF1K \cite{wang2019revisiting}, Hollywood-2 \cite{mathe2014actions}, and UCF-Sports \cite{mathe2014actions}.

\textbf{DHF1K}: This diverse dataset consists of 1,000 videos, annotated by 17 observers while watching videos freely. This dataset allocates 600, 100 and 300 videos as the train set, validation set and test set, respectively. Since the ground truth for the test set is not published publicly, our model’s performance on the test set of this dataset has been evaluated by the DHF1K benchmark team.

\textbf{Hollywood-2}: This large dataset is composed of 1,707 videos extracted from 69
Hollywood movies with 12 action categories. Contrary to the strategy that used by DHF1K for annotation,
Hollywood-2 employed a task-driven viewing approach. The annotations are collected from context recognition (4 observers), free viewing (3 observers) and action recognition (12 observers). We select 823 videos from the train set for training and 884 videos from the test set for testing purpose, removing videos that do not contain enough number of frames for feeding our encoder.

\textbf{UCF Sports}: It contains 150 videos from the UCF sports action dataset, which includes 9 sports classes. Its method of annotation collection is the same as the Hollywood-2, and it also focuses on task-driven viewing. We adopt the same split as \cite{wang2021spatio} with 103 videos for training and 47 videos for testing.

\subsection{Experimental Setup}

Our encoder is initialized with the pretrained weights of the small version of Video Swin Transformer network, namely \textbf{Video Swin-S}, on the Kinetics-400 \cite{DBLP:journals/corr/KayCSZHVVGBNSZ17} dataset. Moreover, the encoder's parameters are updated during our training procedure. We employ the Adam \cite{kingma2014adam} optimizer for training and the initial learning rate and batch size are set to \(10^{-5}\) and 1, respectively. The spatial resolution of the input video frames is set to \(224 \times 384\) and at each iteration, the network receives a video clip with \(T=32\) consecutive frames.

For the train and test procedures, inspired by \cite{zhou2023transformer}, we train our network on the training set of the DHF1K dataset and we adopt early stopping on the validation set of DHF1K. Then, we fine-tuned our pretrained model on the preselected videos from train set of Hollywood-2 and UCF Sports, separately. For inference, the saliency maps for all the video frames are produced based on a sliding window strategy \cite{min2019tased}. In order to generate the saliency map for the initial frames of the clip, the original clip is reversed, so that enough previous temporal context is available.

For evaluating the performance of our model, we employ two types of distribution-based and location-based metrics \citep{bylinskii2018different}. From the former type, we use Similarity Metric (SIM) and Linear Correlation Coefficient (CC) metrics and for the later type we exploit AUC-Judd (AUC-J), Shuffled AUC (S-AUC) and Normalized Scanpath Saliency (NSS) metrics.

\subsection{Result Analysis}

\begin{figure*}[t]
    \centering
    \includegraphics[width=1\textwidth]{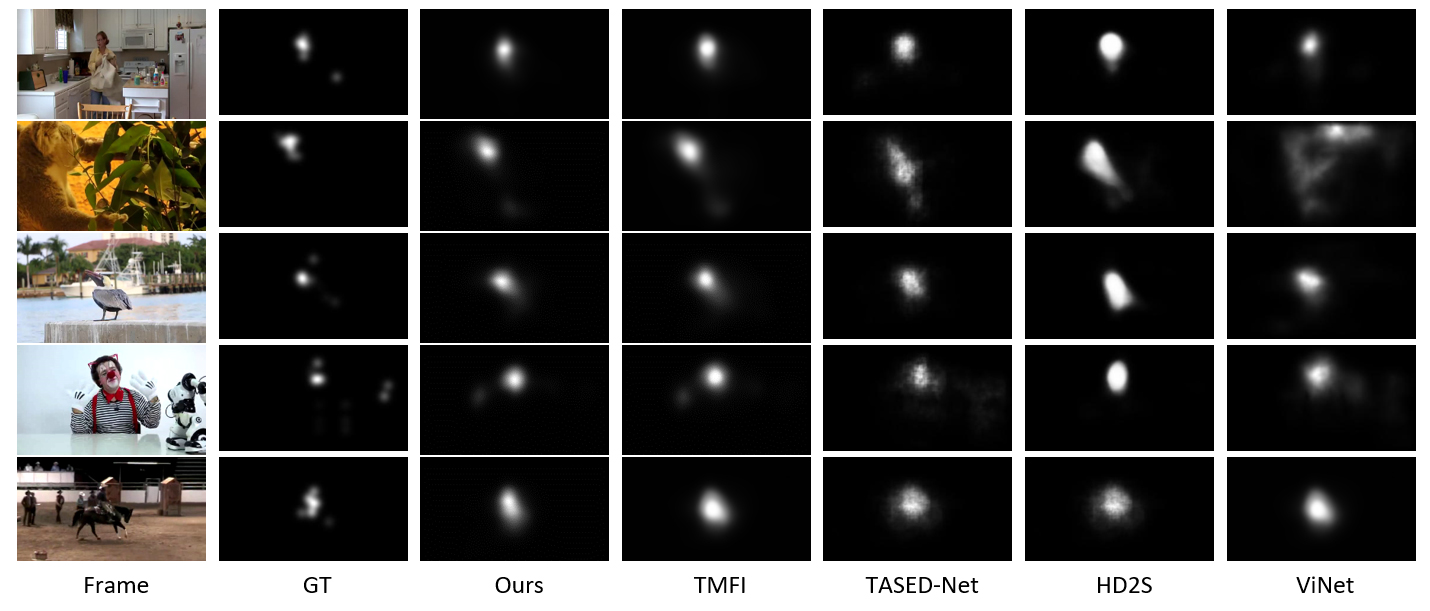}
    \caption{ Qualitative comparison of the performance of different video saliency prediction models.}
    \label{fig2}
\end{figure*}

\begin{table*}
\centering
\rowcolors{2}{gray!25}{white}
\begin{tabular}{|c|c|c|c|c|c|c|c|c|c|c|c|c|c|}
\hline
\rowcolor{gray!50}
\textbf{Models} & \multicolumn{5}{c|}{\textbf{DHF1K}} & \multicolumn{4}{c|}{\textbf{Hollywood-2}} & \multicolumn{4}{c|}{\textbf{UCF Sports}} \\
\hline
& AUC-J & SIM & S-AUC & CC & NSS & AUC-J & SIM & CC & NSS & AUC-J & SIM & CC & NSS \\
\hline
SalEMA & 0.890 & \textbf{0.466} & 0.667 & 0.449 & 2.574 & 0.919 & 0.487 & 0.613 & 3.186 & 0.906 & 0.431 & 0.544 & 2.638 \\
STRA-Net & 0.895 & 0.355 & 0.663 & 0.458 & 2.558 & 0.923 & 0.536 & 0.662 & 3.478 & 0.910 & 0.479 & 0.593 & 3.018 \\
TASED-Net & 0.895 & 0.361 & 0.712 & 0.470 & 2.667 & 0.918 & 0.507 & 0.646 & 3.302 & 0.899 & 0.469 & 0.582 & 2.920 \\
SalSAC & 0.896 & 0.357 & 0.697 & 0.479 & 2.673 & 0.931 & 0.529 & 0.670 & 3.356 & 0.926 & 0.534 & 0.671 & 3.523 \\
UNISAL & 0.901 & 0.390 & 0.691 & 0.490 & 2.776 & 0.934 & 0.542 & 0.673 & 3.901 & 0.918 & 0.523 & 0.644 & 3.381 \\
ViNet & 0.908 & 0.381 & 0.729 & 0.511 & 2.872 & 0.930 & 0.550 & 0.693 & 3.73 & 0.924 & 0.522 & 0.673 & 3.62 \\
HD2S & 0.908 & 0.406 & 0.700 & 0.503 & 2.812 & 0.936 & 0.551 & 0.670 & 3.352 & 0.904 & 0.507 & 0.604 & 3.114 \\
VSFT & 0.9109 & 0.4109 & 0.7200 & 0.5185 & 2.9773 & 0.936 & 0.577 & 0.703 & 3.916 & - & - & - & - \\
TSFP-Net & 0.9116 & 0.3921 & 0.7230 & 0.5168 & 2.9665 & 0.936 & 0.571 & 0.711 & 3.910 & 0.923 & 0.561 & 0.685 & 3.698 \\
STSANet & 0.9125 & 0.3829 & 0.7227 & 0.5288 & 3.0103 & 0.938 & 0.579 & 0.721 & 3.927 & \textbf{0.936} & 0.560 & 0.705 & \textbf{3.908} \\
TMFI-Net & \textbf{0.9153} & 0.4068 & \textbf{0.7306} & 0.5461 & \textbf{3.1463} & \textbf{0.940} & \textbf{0.607} & \textbf{0.739} & \textbf{4.095} & \textbf{0.936} & \textbf{0.565} & 0.707 & 3.863 \\
\textbf{THTD-Net (Ours)} & 0.9152 & 0.4062 & 0.7296 & \textbf{0.5479} & 3.1385 & 0.939 & 0.585 & 0.726 & 3.965 & 0.933 & \textbf{0.565} & \textbf{0.711} & 3.840 \\
\hline
\end{tabular}
\caption{Quantitative comparison of different models on DHF1K, Hollywood-2, and UCF Sports datasets. The top score in each metric is emboldened.}
\label{tab:results}
\end{table*}

We compare the quantitative performance of THTD-Net with top 11 state-of-the-art VSP models that achieved high scores on the DHF1K benchmark\footnote{https://mmcheng.net/videosal}, including SalEMA \cite{linardos2019simple}, STRA-Net \cite{lai2019video}, TASED-Net\cite{min2019tased}, SalSAC \cite{wu2020salsac}, UNISAL \cite{droste2020unified}, ViNet \cite{jain2021vinet}, HD2S \cite{HD2S}, VSFT \cite{DBLP:journals/tcsv/MaSRZL22}, TSFP-Net \cite{chang2021temporal}, STSANet \cite{wang2021spatio}, TMFI-Net \cite{zhou2023transformer}. All methods are evaluated on DHF1K, UCF Sports and Hollywood-2. The quantitative results are reported in Table 1.

\begin{table}    
\centering
\footnotesize{
\rowcolors{2}{gray!25}{white}
\begin{tabular}{|l|c|c|c|c|}
\hline
\rowcolor{gray!50}
     \textbf{Model} & \textbf{Model Size (MB)} \\
\hline
    SalEMA & 364\\
    STRA-Net &641 \\
    TASED-Net &82 \\ 
    SalSAC & 93.5\\ 
    UNISAL &15.5 \\ 
    ViNet &124 \\ 
    HD2S &116 \\ 
    VSFT &71.4 \\ 
    TSPF-Net &58.4 \\ 
    STSANet &643 \\ 
    TMFI-Net &234 \\ \hline
    \textbf{THTD-Net (Ours)} &\textbf{220} \\  
    \hline
\end{tabular}
\caption{Quantitative comparison of the model size (MB).} 
\label{tab:model_size}
}
\end{table}

\begin{table}    
\scriptsize{
\centering
\rowcolors{2}{gray!25}{white}
\begin{tabular}{|l|c|c|c|c|}
\hline
\rowcolor{gray!50}
\textbf{Model} & \textbf{CC} & \textbf{NSS} & \textbf{SIM} & \textbf{AUC-J}\\
\hline
Decoder: 4 layers & 0.551 & 3.168 & 0.419 & 0.923 \\
Decoder: 3 layers & 0.551 & 3.124 & 0.410 & 0.922 \\ 
Decoder: 2 layers & 0.549 & 3.169 & 0.421 & 0.922 \\ 
\hline
Decoder: double layers & 0.553 & 3.169 & 0.414 & 0.923 \\ 
Decoder: triple layers & 0.552 & 3.170 & 0.418 & 0.923 \\ 
\hline
Decoder: MobileNet & 0.545 & 3.119 & 0.416 & 0.921 \\ 
\hline
Decoder: half temporal res. & 0.410 & 2.212 & 0.299 & 0.887 \\ 
\hline
\textbf{THTD-Net (Ours)} & \textbf{0.553} & \textbf{3.188} & \textbf{0.425} & \textbf{0.924} \\
\hline
\end{tabular}
\renewcommand\thetable{2}
\caption{Quantitative comparison of various strategies for decoding temporal features in the decoding stage.}
\label{tab:ablation}}
\end{table}

Based on the results shown in the Table~\ref{tab:results}, on DHF1K dataset, THTD-Net achieves performance that is substantially on par with TMFI-Net and significantly superior to other state-of-the-art methods on metrics such as CC and NSS. 
On the Hollywood-2 and UCF Sports datasets, our model achieves similar results to the best methods of the state of the art, although it does not particularly stand out in the metrics. It should be noted, however, that both Hollywood-2 and UCF Sports address more specific and narrow domains that DHF1K and thus represent an easier task, as also shown by the average higher scores obtained by state-of-the-art methods on all metrics, leading to a flattening of model performance over the various metrics. 
A qualitative comparison of predicted saliency maps on DHF1K samples is illustrated in Fig.~\ref{fig2}, showing how THTD-Net provides consistent and on-target saliency estimation, in spite of lacking complex attention mechanisms and not employing a multi-branch decoding strategy. This demonstrates the validity of the intuition that more effective temporal modeling is a key factor for achieving good performance in video saliency prediction. 

An analysis of model complexity, in terms of number of parameters, is presented in Table~\ref{tab:model_size}. It is interesting to note that our model includes parameters compared to the best two models of the DHF1K benchmark, namely, \cite{zhou2023transformer} and \cite{wang2021spatio}, while providing better performance than the latter and comparable results to the former. Again, this observation emphasizes the effectiveness of the proposed strategy for decoding the spatio-temporal features with high temporal dimensionality, which compensates the requirement on complex feature interactions.

\subsection{Ablation Study}

In this section, with the aim of better assessing the impact of the design choices in the proposed architecture, we present various model variants and evaluate their performance on DHF1K (using the labeled validation set as a test set), showing the results in Table~\ref{tab:ablation}.

In a first set of experiments, we investigate the impact of the number of the decoding layers on the performance of the model. To this aim, we assess how the model performs with a significantly lower number of decoder layers (from 2 to 4, while proportionally changing feature spatio-temporal dimensionality) and, vice versa, with a larger number of decoding layers (twice or three times larger than the baseline architecture, by respectively adding two or three 3D convolutional layer after each base layer, such that newly-added layers do not alter the size of the feature maps). Results show that reducing the number of decoder layers leads to a drop in performance, which is especially noticeable in the NSS metric, while increasing the number of layers does not provide any advantage and even it also leads to slightly worse performance.

We also assess the impact of the choice of simple 3D convolutions in the model's decoder, and compare the performance to a variant which employ resource-efficient 3D-MobileNet blocks \citep{kopuklu2019resource} as a decoder layer. Since each MobileNet block is composed of a depth-wise and a point-wise 3D convolutional layer, we configure the kernel size and stride in such a way as to divide the temporal dimension by a factor of 2 where appropriate. 
Although the number of 3D convolutional layers is increased in this strategy, the number of the parameters of the model is less than THTD-Net, due to separable convolutions. Results, however, exhibit significantly worse performance, hinting that separating the spatial and temporal processing stages in the decoder ends up losing important information for saliency prediction.

Finally, in order to investigate the effectiveness of decoding features with lower temporal dimension, we reduce the temporal dimension of the features by a factor of 2 before feeding them to the decoder of THTD-Net. As per our initial hypothesis, reducing the temporal dimension critically affects the model's capability to effectively predict saliency maps. 

\section{CONCLUSION}
In this paper a transformer-based video saliency prediction model is introduced. The key idea of our approach is to design a model that benefits from the whole temporal information that is provided by the encoder in the decoding stage. This, we avoid reducing the temporal dimension of the encoder's features when feeding them to the decoder, and also ensure that temporal information is reduced gradually within different decoding stages. The model is evaluated on three common benchmarks and the results demonstrate that its performance is comparable to state-of-the-art methods, even though our model is not equipped with attention modules or complex feature interaction mechanisms, and is more parameter-efficient than the currently best-performing model on the DHF1K benchmark, namely, TMFI-Net~\cite{zhou2023transformer}. Our ablation studies confirm that, in order to effectively process higher-resolution temporal features, it is of high importance to employ comparably longer decoder architectures. Nonetheless, at the same time, increasing the number of the decoding layers not only would not improve the performance, but also will increase the number of the model parameters and consequently it may lead to overfitting.

\section*{ACKNOWLEDGMENTS}

The work of Simone Palazzo, who has contributed to the design of the model architecture and the evaluation procedure, has been supported by MUR in the framework of PNRR Mission 4, Component 2, Investment 1.1, PRIN, under project RESILIENT, CUP E53D23016360001.

\bibliographystyle{apalike}
{\small
\bibliography{Ref}}

\end{document}